\documentclass{article}

\usepackage{arxiv}

\usepackage[utf8]{inputenc} 
\usepackage[T1]{fontenc}    
\usepackage{hyperref}       
\usepackage{url}            
\usepackage{booktabs}       
\usepackage{amsfonts}       
\usepackage{nicefrac}       
\usepackage{microtype}      
\usepackage{lipsum}
\usepackage{graphicx}
\graphicspath{ {./images/} }

\usepackage{authblk}
\usepackage{natbib}
\usepackage{subcaption}
\usepackage{tikz}
\usepackage{amsmath}
\usepackage[flushleft]{threeparttable}
\usepackage{multicol}
\usepackage{pgfplots}
\pgfplotsset{compat=1.18} 
\usepackage{bm}
\usepackage{algorithm}
\usepackage{algorithmic}

\title{Deep Reinforcement Learning \\ for Picker Routing Problem in Warehousing}

\author[1]{George Dunn}
\author[2]{Hadi Charkhgard}
\author[3, 4]{Ali Eshragh}
\author[2]{Sasan Mahmoudinazlou}
\author[1]{Elizabeth Stojanovski}
\affil[1]{School of Information and Physical Sciences, University of Newcastle, NSW, Australia}
\affil[2]{Department of Industrial and Management Systems Engineering, University of South Florida, FL, USA}
\affil[3]{Carey Business School, Johns Hopkins University, MD, USA}
\affil[4]{International Computer Science Institute, University of California at Berkeley, CA, USA}

\date{}

\begin{document}
\maketitle
\begin{abstract}
Order Picker Routing is a critical issue in Warehouse Operations 
Management.
Due to the complexity of the problem and the need for quick
solutions, suboptimal algorithms are frequently employed
in practice.
However, Reinforcement Learning offers an appealing alternative
to traditional heuristics, potentially outperforming existing methods
in terms of speed and accuracy.
We introduce an attention based neural network for modeling
picker tours, which is trained using Reinforcement Learning.
Our method is evaluated against existing heuristics
across a range of problem parameters to demonstrate its
efficacy.
A key advantage of our proposed method is its ability to offer an option
to reduce the perceived complexity of routes.
\end{abstract}



\section{Introduction}
\label{introduction}


Warehouse Operations Management encompasses various tasks,
including receiving, storing, order picking, and shipping goods
\cite{rasmi2022wave}.
Order picking involves collecting items listed for retrieval in a pick list from
warehouse locations to meet customer requirements
and transporting them to a depot for packing and shipping
\cite{petersen1997evaluation, de2007design}. 
This process has often been identified as the most expensive task and
can account for up to 55\% of total operating costs
\cite{de2007design, tompkins2010facilities}. 
Due to its significant expense, 
optimizing order efficiency is crucial; therefore,
picker routing policies are implemented to sequence pick
list order to ensure an efficient route through the warehouse.


The task of finding the optimal picker route falls under the category of
Combinatorial Optimization Problems (COPs),
which involves determining
the optimal solution from a discrete but extensive set of possibilities. 
More specifically, it can be viewed as a variant of the Traveling Salesman
Problem (TSP) \cite{scholz2016new}; 
an NP-hard problem that is fundamental in Combinatorial Optimization.
Various algorithms have been developed to obtain optimal solutions
for the order picking problem, 
specifically by leveraging the physical characteristics
of rectangular warehouses. 
However, these algorithms encounter difficulties with scaling 
to handle realistic sized problems with larger volumes and the additional
complexity introduced by side constraints
\cite{ratliff1983order, pansart2018exact}.
In practice, heuristic routing methods are often employed
to obtain simple routes that are easy to follow. 
However, these methods do not guarantee the lowest cost solutions \cite{koster1998routing}.
Studies have shown that even the most effective heuristics produced routes that
were, on average, 9\% larger or worse than the optimal solution
\cite{petersen1997evaluation}. 
Given that order picking is a significant cost, 
it becomes imperative for routing policies to approach
optimal efficiency.

Machine learning techniques represent a potential alternative approach.
By autonomously learning heuristics, machine learning techniques can be
applicable across a wider range of problems without needing to
be specifically developed for each different problem. 
The Reinforcement Learning (RL) subcategory of machine learning is of particular interest,
where an agent can explore the space of possible solutions
to find heuristics without external guidance.
This is beneficial as supervised learning is not
suitable for large COPs due to the impracticability of achieving extensive optimal results.
In the literature, frameworks have been
developed to use RL to solve COPs,
as is evident from recent surveys such as that by \citet{mazyavkina2021reinforcement}. 
Although these studies primarily focus on classical COPs
such as the Traveling Salesman problem (TSP), some,
like those conducted by \citet{nazari2018reinforcement},
extend these frameworks to address other problems such as Vehicle Routing.
However, there remains significant potential to further develop these
methods for wider applications.
Within the warehousing domain, there are very few studies on
RL, and to the best of our knowledge,
none that address picker routing specifically.
The promising results from applying RL to other
COPs suggests that similar
techniques could be successful in addressing the picker routing problem,
and forms the focus of the present research.

In this paper, we develop a model to solve the order
picker routing problem that can be trained with
RL.
Our main contributions can be summarized as follows. 
1) We formulate the method of constructing tour graphs
presented in \citet{ratliff1983order} as a
Markov Decision Process (MDP).
2) We present a Neural Network architecture to model
the MDP formulation based on attention mechanisms and the Transformer
architecture \cite{vaswani2017attention}.
3) The model is modified to produce simplified
tours with reduced complexity for implementation in
human-operated warehouses.
4) Both models are evaluated for a variety of picker routing problems,
providing a comprehensive comparison with existing heuristics.



\section{Related Work}
\label{related}


\textbf{Neural Network Architectures.}
There currently exist many neural network (NN)
architectures developed to model COPs.
The Sequence-to-Sequence model (Seq2seq), first introduced in
\citet{sutskever2014sequence}, uses two separate Recurrent Neural Networks
(RNNs): one to encode the input into an embedding and the other
to decode, by mapping the embedding to an output sequence.
This approach has some limitations for COP problems, as it requires a
fixed output size, hence requiring a new model each time the problem parameters change.
Furthermore, the decoder relies on information contained in a single hidden state
at the end of the input sequence, making it difficult for the model to
propagate enough information to the output for the states early in the input
sequence.
The Attention Model (AM) \cite{bahdanau2014neural} addresses the latter issue
by using another NN to transfer additional contextual information to the output
over the entire sequence of encoder states, allowing the model to learn to
focus specifically on the most relevant entries in the input sequence at
each time step.
The Pointer Network (PN) was introduced by \citet{vinyals2015pointer},
and extends on the AM to generate a permutation of the input.
While the AM processes inputs, the PN directly selects a member of the input
sequence at each step of the output, which is promising for COP problems
with varying output dictionaries, where the outputs correspond to positions
in the input. 
The PN architecture was applied to computing planar convex hulls,
Delaunay triangulation, and the TSP and further demonstrated that the models
generalize beyond the maximum lengths on which they were trained.
\citet{vaswani2017attention} goes on to develop the Transformer model,
which is an encoder-decoder architecture based entirely on attention.
The use of self-attention captures long-term dependencies,
improving the vanishing gradient issues of traditional RNNs.
With a sequence of items as input, the goal of self-attention is to
estimate the relevance of each item to all other items.
This method is built on to provide Multi-Head Attention by linearly
projecting the inputs multiple times and performing the attention
on each projection, enabling simultaneous modeling of different
information.


\textbf{Reinforcement Learning on Combinatorial Optimisation Problems.}
Within the various architectures, frameworks have been developed to  
utilise RL in training these models to solve COPs. 
Using the architecture of the PN model, \citet{bello2016neural}
implement an Actor-Critic algorithm to produce near-optimal solutions for
the TSP problem using the negative tour length as a reward signal.
Therefore, the model parameters are trained with this method,
without the need to use optimal solutions as supervised examples.
This method improved on the method of \citet{vinyals2015pointer}
for large TSPs, with close to optimal results obtained for
TSPs with 100 nodes.
\citet{nazari2018reinforcement} extend on this approach but generalize the
architecture to accommodate a larger range of COPs.
Unlike problems such as Natural Language Processing, the order in which
elements are input into the model holds no significance for COPs. 
By replacing the LSTM encoder of the Pointer Network with input embeddings,
the model becomes invariant to the input sequence.
This adaptation allows the model to be used on a dynamic Vehicle
Routing Problem where model parameters are trained using the
REINFORCE policy gradient method.
Similarly, \citet{deudon2018learning} also extends on the results of
\citet{bello2016neural} to present a model to address TSPs using
attention mechanisms by replacing the LSTM units of the encoder to
ensure that the input is encoded as a set instead of a sequence.
In this study, the parameters were trained using REINFORCE with a
critic baseline.
After the success of the Transformer model
\cite{vaswani2017attention} in other domains,
\citet{kool2018attention} used the architecture to
solve routing problems by proposing a model based on attention layers
with benefits over the pointer network.
The model is trained using REINFORCE with a simple baseline based on
deterministic greedy rollout.
The application of this model on TSP, VRP, and the Orienteering Problem (OP)
outperformed a wide range of existing heuristics and
methods, approaching the optimal results for TSPs up to 100 nodes.
The Graph Pointer Network (GPN) was introduced by \citet{ma2019combinatorial} to
solve TSPs, and builds on PNs by introducing a graph embedding layer
on the input to capture the relationship between the nodes.
This architecture is also trained using a policy gradient.
An in-depth review of the use of RL to solve COPs is
presented in \citet{mazyavkina2021reinforcement}.


\textbf{Reinforcement Learning on Warehousing.}
\citet{cals2020solving} present Deep Reinforcement Learning (DRL) methods to 
decide how and when orders should be batched and picked in a warehouse 
setting, with the goal to minimise the number of tardy orders.
Using a policy gradient method, this was the first
application of DRL to solve the Order Picking
and Sequencing Problem (OBSP),
and the first use of DRL in the warehouse domain.
This method is extended further by \citet{beeks2022deep} to minimize picking costs
by optimizing the trade-off between the short lead time of small
batches and the order picking efficiency of larger batches.
These methods do not consider the exact location of items and instead
use simulated transportation times to decide on picking
methods, and so are not applicable to problems
of finding the shortest path of a picker through a
workshop.
Other methods for applying RL to warehouse management are provided
by \citet{li2019task} and \citet{hu2020deep};
both use DQN models for order assignment problems without
addressing the optimal picker routing problem.


With the limited existing literature on RL in warehousing,
particularly on picker routing, there are many opportunities for
new developments in this area. 
Studies on RL in classical COP problems such as TSP
indicate the potential for similar techniques to produce valuable
results for picker routing problems.


\section{Background}
\label{background}

\begin{figure}[ht]
\vskip 0.2in
\begin{center}
\centering
    \begin{subfigure}[b]{0.34\textwidth}
        \centering
        \resizebox{\linewidth}{!}{
            \begin{tikzpicture}[shorten >=1pt,draw=black!50]

    \draw[black, thin] (0, -1) -- (0 , 5.5) -- (7.5, 5.5) -- (7.5, -0.5)
    -- (1.5, -0.5) -- (1.5, -1) -- cycle;

    \foreach \name / \x in {0,...,4}{
        \draw[black, thin] (1.5 * \x, 0) -- (1.5 * \x , 5);
        \draw[black, thin] (1.5 * \x + 0.5, 0) -- (1.5 * \x + 0.5 , 5);
        \draw[black, thin] (1.5 * \x + 1, 0) -- (1.5 * \x + 1 , 5);
        \draw[black, thin] (1.5 * \x + 1.5, 0) -- (1.5 * \x + 1.5, 5);
        \foreach \name / \y in {0,...,10}{
            \draw[black, thin] (1.5 * \x, 0.5 * \y) -- (1.5 * \x + 0.5, 0.5 * \y);
            \draw[black, thin] (1.5 * \x + 1, 0.5 * \y) -- (1.5 * \x + 1.5, 0.5 * \y);
            }
    }

    \node[align=center] at (0.75, -0.75) 
    {Depot};

    \node[align=center] at (3.75, -0.25) {Front Cross-Aisle};
    \node[align=center] at (3.75, 5.25) {Back Cross-Aisle};
    \node[align=center, rotate=90] at (3.75, 2.5) {Pick-Aisle};


    \node[circle,fill=black,minimum size=3pt] (item1) at (0.25, 1.25) {};

    \node[circle,fill=black,minimum size=3pt] (item2) at (1.75, 4.25) {};

    \node[circle,fill=black,minimum size=3pt] (item3) at (2.75, 1.25) {};

    \node[circle,fill=black,minimum size=3pt] (item4) at (4.75, 3.75) {};

    \node[circle,fill=black,minimum size=3pt] (item5) at (5.75, 3.25) {};

    \node[circle,fill=black,minimum size=3pt] (item6) at (6.25, 0.75) {};

    \node[circle,fill=black,minimum size=3pt] (item7) at (7.25, 4.75) {};

\end{tikzpicture}
        }
        \caption{Warehouse}
        \label{fig:warehouse_fig}
    \end{subfigure}
    \begin{subfigure}[b]{0.35\textwidth}
    \centering
        \resizebox{\linewidth}{!}{
            \begin{tikzpicture}[shorten >=1pt,draw=black!50]


    \draw[black, thin] (0, -0.25) -- (6, -0.25);

    \draw[black, thin] (0, 5.25) -- (6, 5.25);

    \foreach \name / \y in {0,...,4}{
        \node[shape=circle,draw=black, minimum size=3pt, fill=white] (A-\name) at (1.5 * \y, -0.25) {};
        \node[shape=circle,draw=black, minimum size=3pt, fill=white] (B-\name) at (1.5 * \y , 5.25) {};

        \draw[black, thin] (A-\name) -- (B-\name);
        }


    \node[circle, draw=black, minimum size=3pt, fill=black] (depot) at (0, -0.25) {};

    \node[circle, draw=black, minimum size=3pt, fill=black] (item1) at (0, 1.25) {};

    \node[circle,draw=black, minimum size=3pt, fill=black] (item2) at (1.5, 4.25) {};

    \node[circle,draw=black, minimum size=3pt, fill=black] (item3) at (1.5, 1.25) {};

    \node[circle,draw=black, minimum size=3pt, fill=black] (item4) at (4.5, 3.75) {};

    \node[circle,draw=black, minimum size=3pt, fill=black] (item5) at (4.5, 3.25) {};

    \node[circle,draw=black, minimum size=3pt, fill=black] (item6) at (6, 0.75) {};

    \node[circle,draw=black, minimum size=3pt, fill=black] (item7) at (6, 4.75) {};



\end{tikzpicture}
        }
        \caption{Graph Representation}   
        \label{fig:graph}
    \end{subfigure}
\caption{Stardard Rectangular Warehouse} 
\label{fig:warehouse}
\end{center}
\vskip -0.2in
\end{figure}
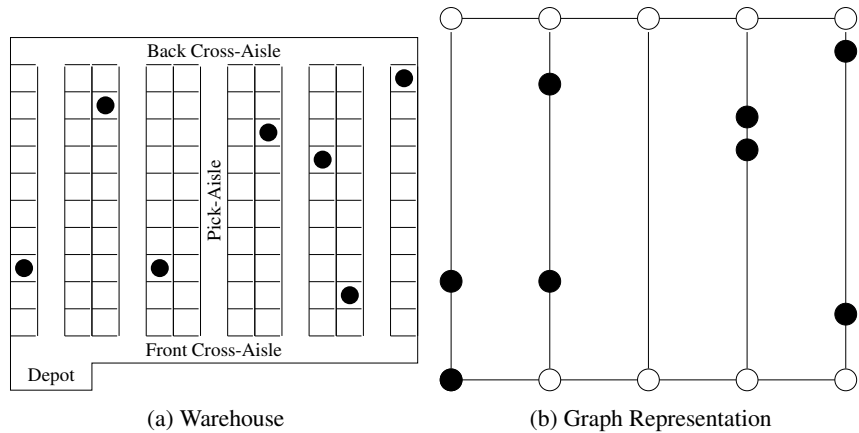

This paper considers a basic rectangular warehouse with a block layout,
a standard configuration used in practice \cite{roodbergen2001routing}. 
The layout consists of multiple vertical aisles in parallel with items
stored on both sides of the aisle, 
two horizontal cross-aisles (one top and one bottom) to allow pickers to
navigate between aisles,
and a single depot as shown in Figure \ref{fig:warehouse}.
We assume that the width of an aisle is negligible; 
therefore, there is no need to consider horizontal distances within an
aisle.

\subsection{Travelling Salesman Problem Formulation}

The Single Picker Routing Problem (SPRP) can
be viewed as a variation of the TSP,
where the items and the depot are the nodes.
A problem instance $L = \{l_0,...,l_m\}$ is defined as a list of locations
within a warehouse, where $l_i = (x_i,y_i) \in \mathbb{R}^2$ denotes the
horizontal and vertical coordinates, respectively.
Let $l_0$ be the depot and $\{l_1,..,l_m\}$
the items in a pick-list of size $m$.
A picker routing tour $\boldsymbol{\pi} = (\pi_0,...,\pi_m)$ 
is a permutation of the items in the given pick list where
$\pi_t \in s$ and $\pi_t \neq \pi_{t'}$ $\forall t \neq t'$.
Valid tours start at the depot and visit each item exactly once
before returning to the depot.
SPRPs can also be formulated as a Steiner TSP
where the items are nodes that can be visited multiple times
and the top and bottom of the aisles are nodes that are not required
to be visited \cite{de2007design}, however,
to be consistent with existing results in DRL,
we present the standard TSP formulation.


\textbf{State.} 
At time-step $t$,
the state $\mathbf{s}_t$ is a partial tour given by
the sequence of all previously selected actions
$\mathbf{\pi}_{0:t-1}$.


\textbf{Action.} 
Defined as selecting one of the nodes that have not yet been selected
$\pi_t \in \{L \setminus \mathbf{\pi}_{1:t-1}\}$.


\textbf{Cost.} 
The cost of selecting the node $\pi_t$ in
state $\mathbf{\pi}_{0:t-1}$ is the negative distance between the nodes.
A traditional TSP considers the Euclidean distance between points;
however, in this order picker routing variation the distance
between two items is a combination of vertical movement
within the aisle along horizontal travel at the cross aisles.
For items in different aisles, the picker could travel
through the top or bottom cross aisle, but
as we are interested in optimizing the distances traveled,
the shortest path is used. 
Let the lower left and upper left corners of a warehouse
be given by the coordinates $b_0 = (0,0)$ and $a_0 = (0,h)$,
respectively,
where $h$ is the length of each aisle.
The distance between two picking positions 
$l_i = (x_i, y_i)$ and $l_j = (x_j, y_j)$
is then calculated as:
\begin{equation}
    d(l_i, l_j) = 
        \begin{cases}
            |y_i-y_j| & \text{if} \ x_i = x_j, \\
            |x_i - x_j| + \min((y_i+y_j),(2h-(y_i+y_j))) & \text{otherwise}.
        \end{cases}
\end{equation}

The total reward for a complete solution is given by the negative tour
length and the goal is to find a minimal length valid tour where the 
objective function is given by:
\begin{equation}
    \boldsymbol{\pi}^* = 
    \operatornamewithlimits{argmin}_{\boldsymbol{\pi}} 
    (\sum_{i=0}^{m-1} d(\pi_i, \pi_{i+1}) \ + \ d(\pi_m, \pi_0)).
\end{equation}


\textbf{Policy.} 
With the nodes selected sequentially until a complete tour is constructed,
the probability of a stochastic policy for selecting a complete tour
$\boldsymbol{\pi}$ for problem instance $L$ can be
factorized using the chain rule as:
\begin{equation}
    p(\boldsymbol{\pi}|L) = \prod_{t=1}^n 
    p(\pi_t|L,\pi_{1:t-1}).
\end{equation}


\subsection{Routing Methods}

\begin{figure}[t]
\vskip 0.2in
\begin{center}
    \begin{subfigure}[b]{0.24\linewidth}
        \centering
        \resizebox{\linewidth}{!}{
        \begin{tikzpicture}[shorten >=1pt,draw=black!50]




    \foreach \name / \y in {0,...,4}{
        \node[shape=circle,draw=black, minimum size=3pt, fill=white] (B-\name) at (1.5 * \y, -0.25) {};
        \node[shape=circle,draw=black, minimum size=3pt, fill=white] (A-\name) at (1.5 * \y , 5.25) {};

        }


    \node[circle, draw=black, minimum size=3pt, fill=black] (depot) at (0, -0.25) {};

    \node[circle, draw=black, minimum size=3pt, fill=black] (item1) at (0, 1.25) {};

    \node[circle,draw=black, minimum size=3pt, fill=black] (item2) at (1.5, 4.25) {};

    \node[circle,draw=black, minimum size=3pt, fill=black] (item3) at (1.5, 1.25) {};

    \node[circle,draw=black, minimum size=3pt, fill=black] (item4) at (4.5, 3.75) {};

    \node[circle,draw=black, minimum size=3pt, fill=black] (item5) at (4.5, 3.25) {};

    \node[circle,draw=black, minimum size=3pt, fill=black] (item6) at (6, 0.75) {};

    \node[circle,draw=black, minimum size=3pt, fill=black] (item7) at (6, 4.75) {};



    \draw[black, thin] (depot) -- (A-0);
    \draw[black, thin] (A-0) -- (A-1);
    \draw[black, thin] (A-1) -- (B-1);
    \draw[black, thin, double, double distance between line centers=5pt] (B-1) -- (B-2);
    \draw[black, thin, double, double distance between line centers=5pt] (B-2) -- (B-3);
    \draw[black, thin] (B-3) -- (A-3);
    \draw[black, thin] (A-3) -- (A-4);
    \draw[black, thin] (A-4) -- (B-4);
    \draw[black, thin] (B-4) -- (B-3);
    \draw[black, thin] (B-1) -- (B-0);

\end{tikzpicture}
        }
        \caption{S-Shape (Traversal)}
        \label{fig:traversal_new}
    \end{subfigure}
    \begin{subfigure}[b]{0.24\linewidth}
        \centering
        \resizebox{\linewidth}{!}{
        \begin{tikzpicture}[shorten >=1pt,draw=black!50]




    \foreach \name / \y in {0,...,4}{
        \node[shape=circle,draw=black, minimum size=3pt, fill=white] (B-\name) at (1.5 * \y, -0.25) {};
        \node[shape=circle,draw=black, minimum size=3pt, fill=white] (A-\name) at (1.5 * \y , 5.25) {};

        }


    \node[circle, draw=black, minimum size=3pt, fill=black] (depot) at (0, -0.25) {};

    \node[circle, draw=black, minimum size=3pt, fill=black] (item1) at (0, 1.25) {};

    \node[circle,draw=black, minimum size=3pt, fill=black] (item2) at (1.5, 4.25) {};

    \node[circle,draw=black, minimum size=3pt, fill=black] (item3) at (1.5, 1.25) {};

    \node[circle,draw=black, minimum size=3pt, fill=black] (item4) at (4.5, 3.75) {};

    \node[circle,draw=black, minimum size=3pt, fill=black] (item5) at (4.5, 3.25) {};

    \node[circle,draw=black, minimum size=3pt, fill=black] (item6) at (6, 0.75) {};

    \node[circle,draw=black, minimum size=3pt, fill=black] (item7) at (6, 4.75) {};



    \draw[black, thin, double, double distance between line centers=5pt] (depot) -- (item1);
    \draw[black, thin, double, double distance between line centers=5pt] (B-1) -- (item3);
    \draw[black, thin, double, double distance between line centers=5pt] (item3) -- (item2);
    \draw[black, thin, double, double distance between line centers=5pt] (B-1) -- (B-2);
    \draw[black, thin, double, double distance between line centers=5pt] (B-2) -- (B-3);
    \draw[black, thin, double, double distance between line centers=5pt] (B-3) -- (item5);
    \draw[black, thin, double, double distance between line centers=5pt] (item5) -- (item4);
    \draw[black, thin, double, double distance between line centers=5pt] (B-4) -- (item6);
    \draw[black, thin, double, double distance between line centers=5pt] (item6) -- (item7);
    \draw[black, thin, double, double distance between line centers=5pt] (B-4) -- (B-3);
    \draw[black, thin, double, double distance between line centers=5pt] (B-1) -- (B-0);

\end{tikzpicture}
        }
        \caption{Return}
        \label{fig:return_new}
    \end{subfigure}
    \begin{subfigure}[b]{0.24\linewidth}
        \centering
        \resizebox{\linewidth}{!}{
        \begin{tikzpicture}[shorten >=1pt,draw=black!50]




    \foreach \name / \y in {0,...,4}{
        \node[shape=circle,draw=black, minimum size=3pt, fill=white] (B-\name) at (1.5 * \y, -0.25) {};
        \node[shape=circle,draw=black, minimum size=3pt, fill=white] (A-\name) at (1.5 * \y , 5.25) {};

        }


    \node[circle, draw=black, minimum size=3pt, fill=black] (depot) at (0, -0.25) {};

    \node[circle, draw=black, minimum size=3pt, fill=black] (item1) at (0, 1.25) {};

    \node[circle,draw=black, minimum size=3pt, fill=black] (item2) at (1.5, 4.25) {};

    \node[circle,draw=black, minimum size=3pt, fill=black] (item3) at (1.5, 1.25) {};

    \node[circle,draw=black, minimum size=3pt, fill=black] (item4) at (4.5, 3.75) {};

    \node[circle,draw=black, minimum size=3pt, fill=black] (item5) at (4.5, 3.25) {};

    \node[circle,draw=black, minimum size=3pt, fill=black] (item6) at (6, 0.75) {};

    \node[circle,draw=black, minimum size=3pt, fill=black] (item7) at (6, 4.75) {};



    \draw[black, thin] (depot) -- (A-0);
    \draw[black, thin] (A-0) -- (A-1);
    \draw[black, thin] (A-1) -- (A-2);
    \draw[black, thin] (A-2) -- (A-3);
    \draw[black, thin] (A-3) -- (A-4);
    \draw[black, thin] (A-4) -- (B-4);
    \draw[black, thin] (B-4) -- (B-3);
    \draw[black, thin] (B-3) -- (B-2);
    \draw[black, thin] (B-2) -- (B-1);
    \draw[black, thin] (B-1) -- (B-0);

    \draw[black, thin, double, double distance between line centers=5pt] (A-1) -- (item2);
    \draw[black, thin, double, double distance between line centers=5pt] (A-3) -- (item4);
    \draw[black, thin, double, double distance between line centers=5pt] (item4) -- (item5);

    \draw[black, thin, double, double distance between line centers=5pt] (B-1) -- (item3);


\end{tikzpicture}
        }
        \caption{Largest Gap}
        \label{fig:largestgap_new}
    \end{subfigure}
    \begin{subfigure}[b]{0.24\linewidth}
        \centering
        \resizebox{\linewidth}{!}{
        \begin{tikzpicture}[shorten >=1pt,draw=black!50]




    \foreach \name / \y in {0,...,4}{
        \node[shape=circle,draw=black, minimum size=3pt, fill=white] (B-\name) at (1.5 * \y, -0.25) {};
        \node[shape=circle,draw=black, minimum size=3pt, fill=white] (A-\name) at (1.5 * \y , 5.25) {};

        }


    \node[circle, draw=black, minimum size=3pt, fill=black] (depot) at (0, -0.25) {};

    \node[circle, draw=black, minimum size=3pt, fill=black] (item1) at (0, 1.25) {};

    \node[circle,draw=black, minimum size=3pt, fill=black] (item2) at (1.5, 4.25) {};

    \node[circle,draw=black, minimum size=3pt, fill=black] (item3) at (1.5, 1.25) {};

    \node[circle,draw=black, minimum size=3pt, fill=black] (item4) at (4.5, 3.75) {};

    \node[circle,draw=black, minimum size=3pt, fill=black] (item5) at (4.5, 3.25) {};

    \node[circle,draw=black, minimum size=3pt, fill=black] (item6) at (6, 0.75) {};

    \node[circle,draw=black, minimum size=3pt, fill=black] (item7) at (6, 4.75) {};



    \draw[black, thin, double, double distance between line centers=5pt] (depot) -- (item1);
    
    \draw[black, thin] (A-1) -- (A-2);
    \draw[black, thin] (A-2) -- (A-3);
    \draw[black, thin] (A-3) -- (A-4);
    \draw[black, thin] (A-4) -- (B-4);
    \draw[black, thin] (B-4) -- (B-3);
    \draw[black, thin] (B-3) -- (B-2);
    \draw[black, thin] (B-2) -- (B-1);
    \draw[black, thin, double, double distance between line centers=5pt] (B-1) -- (B-0);

    \draw[black, thin, double, double distance between line centers=5pt] (A-3) -- (item4);
    \draw[black, thin, double, double distance between line centers=5pt] (item4) -- (item5);

    \draw[black, thin] (B-1) -- (A-1);


\end{tikzpicture}
        }
        \caption{Composite}
        \label{fig:composite_new}
    \end{subfigure}
\caption{Routing Heuristics Tour Graphs} 
\label{fig:heuristics_new}
\end{center}
\vskip -0.2in
\end{figure}
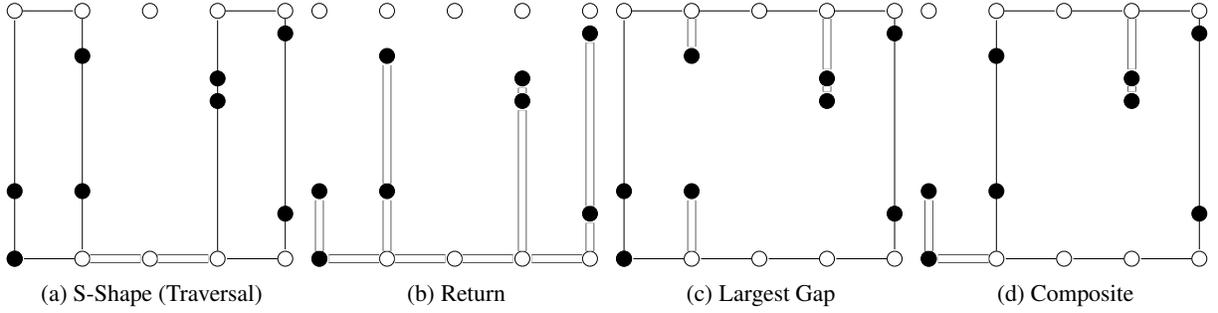

Rule-based heuristic routing methods are commonly used to determine the
sequence of items in a pick-list,
with s-shape, return, midpoint, largest gap, and composite routing policies
being some of the most widely used in literature and practice
\cite{petersen1997evaluation, cano2017evaluation}.
Figure \ref{fig:heuristics_new} shows examples of s-shape,
return, largest gap, and composite routing strategies.
Although the midpoint strategy is also popular in practice,
we do not include it in our comparisons,
as the largest-gap heuristic, by definition, always
produces a solution equal to or better than the midpoint \cite{dukic2007order}.

\textbf{S-Shape Routing Policy.}
The s-shape heuristic, also known as the traversal method,
is one of the most simple routing policies.
It consists of the picker entering each aisle
that contains at least one item to be picked, 
traversing it completely, and
exiting from the other end.
The picker repeats this for all aisles that
contain items before returning to the depot.


\textbf{Return Routing Policy.}
The return routing policy is another of the more simple
strategies.
For each aisle containing a required item,
the picker enters and exits from the front cross-aisle, 
collecting all the items in the pick list.


\textbf{Largest Gap.}
The Largest-Gap heuristic \cite{hall1993distance} 
involves the picker traversing the first and
last aisles containing items completely.
For each aisle in between, the picker enters and exits from the
bottom and top cross-aisles in a way
that all items are visited, and the distance of the aisle that
is not traveled is maximized.


\textbf{Composite Routing Policy.}
The composite routing strategy presented by \citet{petersen1995routeing}
combines the best features of the return and s-shape methods. 
For all adjacent aisles that contain items in a pick list,
the heuristic is chosen to minimize the travel distance between the
farthest picks.


\textbf{Optimal Routing Policy.}
\citet{ratliff1983order} developed an optimal procedure
for picker routing within a rectangular warehouse
that can be solved linearly with respect to the number of aisles
and pick locations.
This algorithm sequentially builds 
Partial Tour Sub-graphs (PTS), where the nodes are made up of the 
top and bottom of each aisle along with the location of each 
item to be picked.
\citet{roodbergen2001routing} extends the algorithm in
\citet{ratliff1983order} for cases with three cross-aisles,
with the solution being polynomial in the number of aisles and items.
\citet{pansart2018exact} extends this method further
to generalize it to any number of cross-aisles,
providing an efficient method for realistic-sized warehouses. 
However, the problem with this method
is that it does not accommodate side constraints.
In contrast, \citet{pansart2018exact} also presents a MILP
formulation that better accommodates side constraints but does not
scale well to realistic-sized problems due to the
complexity of the model.




\section{The Proposed Tour-Graph Markov Decision Process Formulation}
\label{formulation}


There are many ways to formulate and solve the standard TSP
\cite{oncan2009comparative}, however, the problem
is NP-hard, which means that it cannot be solved optimally in
polynomial time \cite{papadimitriou1998combinatorial}.
Furthermore, solving the TSP formulation involves finding a sequence of items
that then requires an additional step to calculate the minimum distance
between each consecutive pair \cite{lu2016algorithm}.
The dynamic programming algorithm presented by
\citet{ratliff1983order} takes advantage of the geometric structure
of a warehouse, resulting in a more efficient method of producing
an optimal route.
We use the results of \citet{ratliff1983order}
to present an MDP that constructs a tour aisle-by-aisle
rather than item-by-item.
A problem instance $L$ is defined the same as in the TSP formulation.



\begin{figure}
\centering
\vskip 0.2in
\begin{center}
    
\def\v{1.5}
\def\h{0.9}

\begin{tikzpicture}[shorten >=1pt,->,draw=black!50, node distance=\layersep]

\tikzset{minimum size=20pt}

    \foreach \name / \y in {1,...,4}{
        \node[shape=circle,draw=black] (A-\name) at (\y * \h, 3 * \v) {$a_i$};
        \node[shape=circle,draw=black] (top-\name) at (\y * \h, 2 * \v) { };
        \node[shape=circle,draw=black] (bottom-\name) at (\y * \h, \v) { };
        \node[shape=circle,draw=black] (B-\name) at (\y * \h, 0) {$b_i$};
    }

    \foreach \name / \y in {0, 1}{
        \node[shape=circle,draw=black] (A11-\name) at (5 * \h + 2.5 * \y * \h, 3 * \v) {$a_i$};
        \node[shape=circle,draw=black] (B11-\name) at (5 * \h + 2.5 * \y * \h, 2 * \v) {$b_i$};
        \node[shape=circle,draw=black] (A12-\name) at (5 * \h + 2.5 * \y * \h, \v) {$a_i$};
        \node[shape=circle,draw=black] (B12-\name) at (5 * \h + 2.5 * \y * \h, 0) {$b_i$};

        \node[shape=circle,draw=black] (A21-\name) at (1.5 * \h + 5 * \h + 2.5 * \y * \h, 3 * \v) { };
        \node[align=center] at (A21-\name) {$a_{i+1}$};
        \node[shape=circle,draw=black] (B21-\name) at (1.5 * \h + 5 * \h + 2.5 * \y * \h, 2 * \v) { };
        \node[align=center] at (B21-\name) {$b_{i+1}$};
        \node[shape=circle,draw=black] (A22-\name) at (1.5 * \h + 5 * \h + 2.5 * \y * \h, \v) { };
        \node[align=center] at (A22-\name) {$a_{i+1}$};
        \node[shape=circle,draw=black] (B22-\name) at (1.5 * \h + 5 * \h + 2.5 * \y * \h, 0) { };
        \node[align=center] at (B22-\name) {$b_{i+1}$};
    }

    
    \draw[-, thick] (A-1) -- (top-1);
    \draw[-, thick] (top-1) -- (bottom-1);
    \draw[-, thick] (bottom-1) -- (B-1);
    
    \draw[-, thick, double, double distance between line centers=5pt] (A-2) -- (top-2);
    \draw[-, thick, double, double distance between line centers=5pt] (top-2) -- (bottom-2);
    
    \draw[-, thick, double, double distance between line centers=5pt] (top-3) -- (bottom-3);
    \draw[-, thick, double, double distance between line centers=5pt] (bottom-3) -- (B-3);
    
    \draw[-, thick, double, double distance between line centers=5pt] (A-4) -- (top-4);
    \draw[-, thick, double, double distance between line centers=5pt] (bottom-4) -- (B-4);
    
    
    \node[below of=B-1, node distance=0.75cm] (i) {(i)};
    \node[below of=i, node distance=0.6cm, rotate=45] {1pass};
    \node[below of=B-2, node distance=0.75cm] (ii) {(ii)};
    \node[below of=ii, node distance=0.6cm, rotate=45] {top};
    \node[below of=B-3, node distance=0.75cm] (iii) {(iii)};
    \node[below of=iii, node distance=0.6cm, rotate=45] {bottom};
    \node[below of=B-4, node distance=0.75cm] (iv) {(iv)};
    \node[below of=iv, node distance=0.6cm, rotate=45] {gap};


    \draw[-, thick] (A11-0) -- (A21-0);
    \draw[-, thick] (B11-0) -- (B21-0);

    \draw[-, thick, double, double distance between line centers=5pt] (B12-0) -- (B22-0);

    \draw[-, thick, double, double distance between line centers=5pt] (A11-1) -- (A21-1);

    \draw[-, thick, double, double distance between line centers=5pt] (A12-1) -- (A22-1);
    \draw[-, thick, double, double distance between line centers=5pt] (B12-1) -- (B22-1);

    \node[below of=B11-0, node distance=0.75cm] (v) { };
    \node[right of=v, node distance=0.75cm * \h] {(v) $11$};

    \node[below of=B12-0, node distance=0.75cm] (vii) { };
    \node[right of=vii, node distance=0.75cm * \h] {(vii) $02$};

    \node[below of=B11-1, node distance=0.75cm] (vi) { };
    \node[right of=vi, node distance=0.75cm * \h] {(vi) $20$};

    \node[below of=B12-1, node distance=0.75cm] (viii) { };
    \node[right of=viii, node distance=0.75cm * \h] {(viii) $22$};

\end{tikzpicture}

\vskip -0.1in

\caption{Possible edges in optimum tour graph where (i)-(iv) are the
vertical edges within aisle $i$ and (v)-(viii) are the horizontal edges 
between aisle $i$ and aisle $i+1$.}
\label{fig:action_new}
\end{center}
\vskip -0.2in
\end{figure}
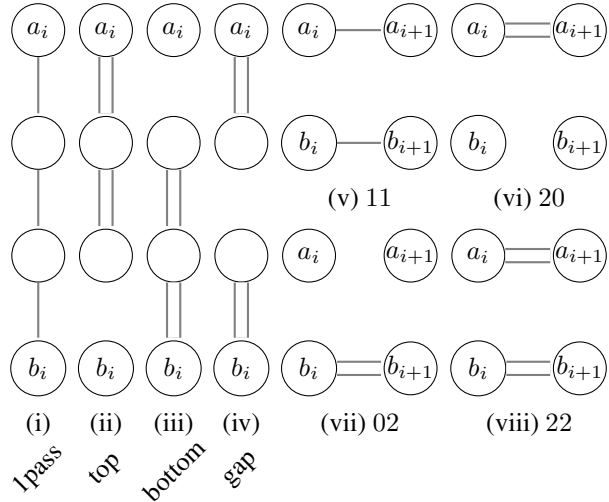

The algorithm of \citet{ratliff1983order} starts in the first
aisle and alternates between actions to decide how a pick-aisle
is traversed (vertical edges) and how cross-aisles are traversed
(horizontal edges) from all aisles in the warehouse.
If empty aisles are ignored and all aisles are of equal length, 
the only possible edge combinations are
shown in Figure \ref{fig:action_new}.
The four options for pick-aisle travel are;
a single traversal of the aisle (1pass),
entering and exiting from the back cross aisle (top),
entering and exiting from the bottom (bottom), 
and traversing the aisle in a way that leaves the largest
gap (gap).
There are also four options available to transition from
aisle $i$ to aisle $i+1$, as shown in
Figure \ref{fig:action_new};
a single pass of the top and bottom (11),
a double pass of the top (20),
a double pass of the bottom (02),
and a double pass of the top and bottom (22).

\textbf{Action.}
The action space $A = A^+ \cup A^-$ for the proposed MDP is made up 
of these edge combinations where:
\begin{align}
    A^+ &= \{\text{1pass, top, bottom, gap}\}, \\
    A^- &= \{11, 20, 02, 22\}.
\end{align}


\textbf{State.}
After each action, a Partial Tour Sub-graph (PTS) in aisle $i$ 
is defined as the graph consisting of all vertices and edges
of aisles 1 to $i$.
\citet{ratliff1983order} define seven states used to characterize
a Partial Tour, which we use as the
state of the MDP, $s_t \in S$, where:
\begin{equation*}
\begin{split}
    S = \{UU1C, 0E1C, E01C, EE1C, EE2C, 000C, 001C\}.
\end{split}
\end{equation*}
The first two symbols are the degree of the rightmost top and bottom
aisle nodes ($0$, uneven $U$, or even $E$ number of edges),
and the last two symbols represent the number of connected components
in the PTS (empty $0C$, one $1C$, or two $2C$).
These were proven to be the only possible states in an optimal tour
graph, and furthermore, it was shown that $001C$ can
be removed if empty aisles are ignored.
Examples of these states are provided in Appendix \ref{sec:pts_states}.


\textbf{Cost.}
The cost of action $a_t \in \mathcal{A}$ in state $s_t$
is the total length of the edges added, $d(s_t, a_t)$.
For aisle $i$ with length $h$,
let $x_i$ be the horizontal coordinate and
$\mathbf{y}_i$ be the set of vertical coordinates for all items
within the aisle.
The costs are then given by:
\begin{align}
    & d(s_t, \text{1pass}) = h, \\
    & d(s_t, \text{top}) = 2(h - \min(\mathbf{y}_i)), \\
    & d(s_t, \text{bottom}) = 2(\max(\mathbf{y}_i)), \\
    & d(s_t, \text{gap}) = 2(h - \text{largestgap}(\mathbf{y}_i)), \\
    & d(s_t, 20) = d(s_t, 02) = d(s_t, 11) = 2(x_{i+1} - x_i), \\
    & d(s_t, 22) = 4(x_{i+1} - x_i).
\end{align}


\textbf{Transition.}
The state transitions are given by the states resulting
from the action taken in the current state.
Tables \ref{tab:aisle_equivalence} and \ref{tab:cross_equivalence}
show the transitions for the pick-aisle and cross-aisle actions,
respectively.
We consider only actions that result in a valid transition for
each equivalence class.
\begin{equation}
    s_{t+1} = T(s_t, a_t).
\end{equation}

\begin{table}[t]
\centering
\caption {Vertical Equivalence Class Transitions $T(s_t, a_t)$}
\label{tab:aisle_equivalence}
\vskip 0.15in
\begin{center}
\begin{small}
\begin{sc}
\begin{threeparttable}

    \begin{tabular}{| c | c  c  c  c  |}
    \hline 
    State & & \multicolumn{2}{c}{Action $a_t$} & \\
    $s_t$ & 1pass & top & bottom & gap  \\
    \hline 
    UU1C & EE1C & UU1C & UU1C & UU1C  \\
    E01C & UU1C & E01C & EE2C & EE2C  \\
    0E1C & UU1C & EE2C & 0E1C & EE2C  \\
    EE1C & UU1C & EE1C & EE1C & EE1C  \\
    EE2C & UU1C & EE2C & EE2C & EE2C  \\
    000C$^a$ & UU1C & E01C & 0E1C & EE2C \\
    \hline
    \end{tabular}
    \begin{tablenotes}
        \small
        \item $^a$ Only possible in initial state.
    \end{tablenotes}
    \vskip -0.1in
\end{threeparttable}
\end{sc}
\end{small}
\end{center}
\end{table}
\begin{table}[t]
\centering
\caption {Horizontal Equivalence Class Transitions $T(s_t, a_t)$}
\label{tab:cross_equivalence}
\vskip 0.15in
\begin{center}
\begin{small}
\begin{sc}
\begin{threeparttable}
    \begin{tabular}{| c | c  c  c  c  |}
    \hline 
    State & & \multicolumn{2}{c}{Action $a_t$} & \\
    $s_t$ & 11 & 20 & 02 & 22 \\
    \hline 
    UU1C & UU1C & $-^a$ & $-^a$ & $-^a$ \\
    E01C & $-^a$ & E01C & $-^b$ & EE2C \\
    0E1C & $-^a$ & $-^b$ & 0E1C & EE2C \\
    EE1C & $-^a$ & E01C & 0E1C & EE1C \\
    EE2C & $-^a$ & $-^b$ & $-^b$ & EE2C \\
    000C$^c$ & $-$ & $-$ & $-$ & $-$ \\
    \hline
    \end{tabular}
    \begin{tablenotes}
        \small
        \item $-$ Never a valid action.
        \item $^a$ The degrees of $a_i$ and $b_i$ are odd.
        \item $^b$ No completion can connect the graph.
        \item $^c$ Only possible in initial state.
    \end{tablenotes}
    \vskip -0.2in
\end{threeparttable}
\end{sc}
\end{small}
\end{center}
\end{table}



\textbf{Policy.}
For a warehouse with $n$ nonempty aisles,
a tour $\boldsymbol{\pi} = (\pi_1,..,\pi_{2n-1})$ 
is a sequence of actions where $\pi_{2i-1} \in \mathcal{A}^+$ 
are vertical edges added within the aisle $i$ 
and $\pi_{2i} \in \mathcal{A}^-$ 
horizontal edges from the aisle $i$ to $i+1$.
The goal is to find a minimal length valid tour where the objective 
function
is given by:
\begin{equation}
    \boldsymbol{\pi}^* = 
    \operatornamewithlimits{argmin}_{\boldsymbol{\pi}} 
    \sum_{i=1}^{2n-1} d(s_i, \pi_i).
\end{equation}

The probability of a stochastic policy for selecting a sequence 
of actions $\boldsymbol{\pi}$ to produce a tour graph 
for problem instance $L$ can be factorised as:
\begin{equation}
    p(\boldsymbol{\pi}|L) = 
    \prod_{t=1}^{2n-1} p(\pi_{t}|s_t).
\end{equation}




\section{Proposed Neural Network Architecture}
\label{model}

\begin{figure}[!ht]
\vskip 0.2in
\centering


\scalebox{0.9}{
\begin{tikzpicture} 
\definecolor{emb_color}{RGB}{252,224,225}
\definecolor{multi_head_attention_color}{RGB}{252,226,187}
\definecolor{add_norm_color}{RGB}{242,243,193}
\definecolor{ff_color}{RGB}{194,232,247}
\definecolor{softmax_color}{RGB}{203,231,207}
\definecolor{linear_color}{RGB}{220,223,240}
\definecolor{mask_color}{RGB}{215,229,235}
\definecolor{gray_bbox_color}{RGB}{243,243,244}

\node[text width=2.5cm, anchor=north, align=center] at (0, 0.5) {Inputs};

\draw[line width=0.05cm] (0.5, 1) -- (3.5, 1) -- (3.5, 1.5) -- (0.5, 1.5) -- cycle;
\node[text width=2.5cm, align=center] at (2, 1.25) {Input Embedding};
\draw[line width=0.05cm, -latex] (0.75, 0.25) -- (2, 0.25) -- (2, 1);

\draw[line width=0.05cm] (-0.5, 1) -- (-3.5, 1) -- (-3.5, 1.5) -- (-0.5, 1.5) -- cycle;
\node[text width=2.5cm, align=center] at (-2, 1.25) {Aisle Encoding};
\draw[line width=0.05cm, -latex] (-0.75, 0.25) -- (-2, 0.25) -- (-2, 1);

\draw[line width=0.05cm, -latex] (-2, 1.5) -- (-2, 2) -- (-0.2, 2);
\draw[line width=0.05cm, -latex] (2, 1.5) -- (2, 2) -- (0.2, 2);
\draw[line width=0.05cm] (0, 2) circle (0.2);
\draw[line width=0.05cm] (-0.2, 2) -- (0.2, 2);
\draw[line width=0.05cm] (0, 1.8) -- (0, 2.2);

\draw[fill=gray_bbox_color, line width=0.05cm] (-2.5, 8.25) -- (2.5, 8.25) -- (2.5, 2.5) -- (-2.5, 2.5) -- cycle;

\draw[line width=0.05cm, fill=white] (-1.5, 3.5) -- (1.5, 3.5) -- (1.5, 4.5) -- (-1.5, 4.5) -- cycle;
\node[text width=2.500000cm, align=center] at (0, 4) {Multi-Head \vspace{-0.05cm} \linebreak Attention};
\draw[line width=0.05cm, -latex] (0, 2.2) -- (0, 3.5);
\draw[line width=0.05cm, -latex] (0, 3) -- (-0.75, 3) -- (-0.75, 3.5);
\draw[line width=0.05cm, -latex] (0, 3) -- (0.75, 3) -- (0.75, 3.5);

\draw[line width=0.05cm, fill=white] (-1.5, 5) -- (1.5, 5) -- (1.5, 5.5) -- (-1.5, 5.5) -- cycle;
\node[text width=2.5cm, align=center] at (0, 5.25) {Add \& Norm};
\draw[line width=0.05cm, -latex] (0, 4.5) -- (0, 5);
\draw[line width=0.05cm, -latex] (0, 2.75) -- (-2, 2.75) -- (-2, 5.25) -- (-1.5, 5.25);

\draw[line width=0.05cm, fill=white] (-1.5, 6.5) -- (-1.5, 7) -- (1.5, 7) -- (1.5, 6.5) -- cycle;
\node[text width=2.500000cm, align=center] at (0, 6.75) {Feed Forward};
\draw[line width=0.05cm, -latex] (0, 5.5) -- (0, 6.5);

\draw[line width=0.05cm, fill=white] (-1.5, 7.5) -- (-1.5, 8) -- (1.5, 8) -- (1.5, 7.5) -- cycle;
\node[text width=2.500000cm, align=center] at (0, 7.75) {Add \& Norm};
\draw[line width=0.05cm, -latex] (0, 7) -- (0, 7.5);
\draw[line width=0.05cm, -latex] (0, 6) -- (-2, 6) -- (-2, 7.75) -- (-1.5, 7.75);

\draw[line width=0.05cm, fill=white] (-1.5, 8.75) -- (-1.5, 9.25) -- (1.5, 9.25) -- (1.5, 8.75) -- cycle;
\node[text width=2.500000cm, align=center] at (0, 9) {Linear};
\draw[line width=0.05cm, -latex] (0, 8) -- (0, 8.75);



\node[text width=2.5cm, anchor=south, align=center] at (0, 9.75) {Outputs};
\draw[line width=0.05cm, -latex] (0, 9.25) -- (0, 9.75);

\node[anchor=east] at (-2.75, 5.25) {$N\times$};

\end{tikzpicture}}
\caption{Proposed Model for Picker Routing Problem}
\label{fig:model}
\end{figure}
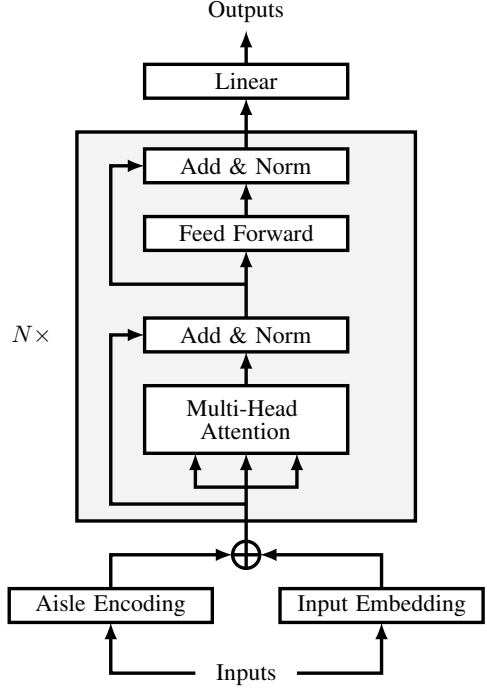

With the Tour Construction MDP presented in the previous section, 
we develop a Neural Network architecture that utilizes attention
mechanisms to approximate the policy.
Previous methods to solve TSPs with RL output nodes
one at a time produce an output that is a permutation of the input
\cite{bello2016neural,nazari2018reinforcement,deudon2018learning,
kool2018attention,ma2019combinatorial}.
Our method sequentially outputs the vertical and horizontal edges
to be added to a tour graph until a complete tour graph is constructed.
The resulting model is shown in Figure \ref{fig:model}.

\subsection{Inputs}

As this method considers aisles rather than individual items,
the inputs are not a sequence of two-dimensional coordinates,
but a sequence of binary vectors representing each aisle.
The size of the sequence is equal to the number of aisles in the warehouse,
and the dimension of each vector is equal to the number of possible storage
locations in each aisle.

\textbf{Input Sequence.}
For a warehouse with $h$ storage locations within each
aisle,
we map the problem instance $L = \{l_0,..,l_m\}$
to the input sequence $\mathbf{z} = \{z_1,...,z_n\}$ where
$n$ is the number of non-empty aisles and
$z_i \in \mathbb{R}^{d_z}$.
The dimensions of the inputs are $d_z = h$ to include all pick
locations.
The $j^{th}$ element of vector $z_i$ is:
\begin{equation}
    z_{ij} = 
        \begin{cases}
            1 & \text{if} \ (i,j) \in L, \\
            0 & \text{else}. \\
        \end{cases}
\end{equation}

\textbf{Input Embedding.}
The inputs $\mathbf{z}$ with dimension $d_z$
go through a learned linear projection to the hidden dimension
$d_h$, through
$\mathbf{h}_i = W^z z_i + \mathbf{b}^z$.
Similarly to \citet{vaswani2017attention},
we multiply these weights by $\sqrt{d_h}$.

\textbf{Aisle Encoding.}
Unlike traditional recurrent models, attention-based encoders are
invariant to the input order.
This was beneficial for the Attention Model in \citet{kool2018attention} 
used to solve TSPs, as the sequence in which the nodes are input
is not relevant.
In the case of our model, the order of the input sequence corresponds
to the horizontal positions of the respective aisles; therefore,
it is important for the model to capture this information.
The Transformer model in \citet{vaswani2017attention} adds an
encoding to the input embeddings to inject information
on the relative position of words within a sentence for its applications
to Natural Language Processing.
Although our class of problem is different,
this method also allowed us to capture the relevant
horizontal information.
Therefore, the ``aisle encodings'' we use are similar to the
``position encodings'' in \citet{vaswani2017attention}, 
defined by sine and cosign functions of different frequencies,
as shown in Appendix \ref{sec:encoding}.


\subsection{Encoder Layers}

Following on from the Transformer encoder in \citet{vaswani2017attention},
we use $N$ encoder layers, each consisting of two sub-layers;
a Multi-Head Attention (MHA) layer and
an aisle-wise fully connected feedforward (FF) layer.
Each sublayer adds a skip connection \cite{he2016deep}
and layer normalization \cite{ba2016layer}.
More detail is provided in Appendix \ref{sec:encoder}.

We implement an attention mask to prevent aisle embeddings
from attending to previous aisles,
as at each step, we only need to consider the current
aisle and all future aisles.
The mask is therefore given by:
\begin{equation}
    M_{ij} = 
    \begin{cases}
        \text{True} & \text{if} \quad i>j, \\
        \text{False} & \text{otherwise}. \\
    \end{cases}
\end{equation}
This also provides our model with the ability to
be applied to batches that contain problems of different
sizes.
By adding padding tokens to the start of the inputs
to make the sequence length the same size across the
whole batch,
the attention mask ensures that the aisles do not
attend to the padded inputs.


\subsection{Output}

A learned linear transformation is used to convert the output of
the last attention layer to the output dimension $d_{out}$.
As we require a probability distribution over the possible
actions,
the dimension size is the maximum number of available actions.
To simplify the model and eliminate the need for two separate
outputs,
we incorporate vertical-horizontal action pairs, where
actions $a = (a^+, a^-)$ for $a^+ \in \mathcal{A}^+$
and $a^- \in \mathcal{A}^-$ give $d_{out} = 16$.
Grouping action pairs also helps the model to capture the
relationships between vertical and horizontal actions.

Following \citet{bello2016neural} and \citet{kool2018attention},
the output is clipped within [-C, C] (C=10) using tanh:
\begin{equation}
    h'_j = C \cdot \text{tanh}(h_j).
\end{equation}


\subsection{Tour Graph Construction}

The output of the model is interpreted as unnormalized log-probabilities
(logits).
This model allows for the logits of all actions in all aisles to be
generated at once with a single forward pass.
The solutions are then constructed sequentially for each aisle, 
calculating the final probabilities $\bm{p}$ using a softmax
after masking non-valid actions.
For aisle $i$,
the output for action $j$ is:
\begin{equation}
    h^{out}_{ij} = 
        \begin{cases}
            h'_{ij} & \text{if action $j$ is valid in state $s_i$}, \\
            -\infty & \text{otherwise}. \\
        \end{cases}
\end{equation}
Valid actions are as follows:
\begin{itemize}
    \item Actions that result in a valid equivalence class. 
    \item The action ``gap'' is only available if the number
    of items in an aisle is strictly greater than 1.
    \item The action taken in the horizontal transition to the final
    aisle must not result in $EE2C$, as there is no way to
    complete this graph.
    \item In the final aisle, actions are only valid if they result
    in equivalences $E01C$, $0E1C$ or $EE1C$ since the graph
    needs to be connected with all cross-aisle nodes having
    even or zero parities to construct a valid tour.
\end{itemize}

The probability of action $j$ in aisle $i$ is then given by:
\begin{equation}
    p_{ij} 
    = p_\theta(\pi_t = j | s_i) 
    = \frac{e^{h^{out}_{ij}}}{\sum_{j'} e^{h^{out}_{ij'}}}.
\end{equation}

Starting in the state $s_0 = 000C$,
an action is selected in each aisle either by sampling or greedy selection,
and the resulting equivalence class is determined.
This is repeated for all nonempty aisles.



\subsection{Simplifying Tour Complexity}
\label{sec:Simplifying}
A key advantage of our proposed approach is its ability
to reduce the perceived complexity of the routes.
In practice, order pickers often prefer simple routes
and more complex tours, such as those that can occur in
optimal solutions and
may contain characteristics that are difficult to understand
and inconsistent \cite{petersen1997evaluation, hall1993distance}.
This can result in additional orientation effort,
reducing picker speed, and increasing the risk of errors
\cite{petersen1999evaluation}.
One characteristic that is considered to produce
complex tours is entering an aisle more than once
\cite{roodbergen2001multiple}.
The ``gap" action in our formulation is the only decision
that produces such a route \cite{scholz2016new},
so excluding this would guarantee a simple solution
that enters each aisle at most once.
Enforcing this in the TSP formulation would be non-trivial,
but our proposed model allows for the incorporation of such
constraints simply by masking these actions when constructing a tour.


\section{Numerical Experiments}
\label{experiments}


\subsection{Instance Generation}

To evaluate our proposed model,
numerical experiments were performed on a variety
of problem classes to compare the results
against the optimal and existing heuristic methods.
The sizes of the warehouse and the pick-lists were
chosen according to
\citet{scholz2016new} and
\citet{henn2012tabu}.
Each vertical aisle consists of a total of 90 storage locations,
with 45 on each side.
The distance between each adjacent picking location is one
length unit (LU).
The top and bottom cross-aisles are 1LU vertically from the top
and bottom pick locations, respectively, and
the horizontal distance between consecutive aisles is 5LU
with the depot located at the front of the first aisle.
The number of aisles $n$ was 5, 10, 15, 20, 25 and 30 and
the size of the pick lists $m$ was 30, 45, 60, 75 and 90,
with the combinations of these problem parameters
producing 30 problem classes. 
Problem instances were generated randomly, 
with item locations normally distributed.


\subsection{Training}

Similarly to \citet{kool2018attention},
the model was trained with gradient descent
using the REINFORCE \cite{williams1992simple} gradient estimator with
a greedy baseline.
However, it was also divided by the baseline loss as a form
of normalisation:
\begin{equation}
    \nabla \mathcal{L}(\theta|s) = \mathbf{E}_{p_\theta(\boldsymbol{\pi}|s)}
    [\frac{(L(\boldsymbol{\pi}) - b(s))}{b(s)} \nabla
    \log p_\theta(\boldsymbol{\pi}|s)],
\end{equation}
for instance $s$, where $\theta$ is the learnable parameters of the model and the
probability distribution $p_\theta(\boldsymbol{\pi}|s)$ is the output.
The loss
$\mathcal{L}(\theta|s) = \mathbf{E}_{p_\theta(\boldsymbol{\pi}|s)}[L(\boldsymbol{\pi})]$
is the length of the pick route.
The baseline $b(s)$ is given by a greedy deterministic rollout
of the best model so far and is updated as better models
are found.

The loss used in \citet{kool2018attention} is divided
by the baseline cost as the tour lengths for this 
problem are larger than a TSP within a unit square. 
This adaptation also allowed better generalization of
the loss across different sized problems,
as our applications involve different warehouse
and pick-list sizes, where the resulting tour lengths
can vary significantly.
The network is trained using Adam \cite{kingma2014adam} as the optimizer algorithm,
as shown in Appendix \ref{sec:reinforce}.

Both models used a hidden dimension size $d_h = 128$, 
$H = 8$ attention heads, $N = 3$ encoder layers,
and learning rate $\mu = 10^{-5}$.
The standard model was trained for 100 epochs
on 100 batches of 16 instances from all 30 problem classes.
We found that smaller batch sizes worked better for our model,
consistent with the results of \citet{masters2018revisiting}
that found that small batch training achieved the best
stability and generalization performance.
For the simplified tour model, training on larger
warehouse sizes only produced better results as
without the largest gap action available,
the performance of s-shape type routes resulted in a local
optima in problems with fewer aisles.
This model was trained on 200 batches of 16 instances
for all problem classes with 25 or 30 aisles for 150
epochs;
keeping the total number of training steps
equal to that of the standard model.


\subsection{Results}

The evaluation of the proposed model was performed on 100
examples from each problem class, with the results shown
in Appendix \ref{sec:results_table}.
Figure \ref{fig:results} demonstrates the average
optimality gap of the proposed model for
standard and simplified tour constructions
(see Section \ref{sec:Simplifying}).
Comparisons with s-shape, return, composite,
and largest gap routing strategies show
that both proposed methods consistently
outperform heuristics,
with the standard routing construction performing
better than the simplified method.
Our proposed methods also show improved generalization characteristics;
performing consistently over the entire range problem
classes, as opposed to the other methods that clearly
demonstrate better performance on certain problem
sizes.

\begin{figure}[t]
\vskip 0.2in
\centering
    \begin{subfigure}[b]{0.45\textwidth}
        \centering
        \resizebox{\linewidth}{!}{
\begin{tikzpicture}

\definecolor{crimson2143940}{RGB}{214,39,40}
\definecolor{darkgray176}{RGB}{176,176,176}
\definecolor{darkorange25512714}{RGB}{255,127,14}
\definecolor{forestgreen4416044}{RGB}{44,160,44}
\definecolor{lightgray204}{RGB}{204,204,204}
\definecolor{mediumpurple148103189}{RGB}{148,103,189}
\definecolor{sienna1408675}{RGB}{140,86,75}
\definecolor{steelblue31119180}{RGB}{31,119,180}

\begin{axis}[
legend cell align={left},
legend style={
  fill opacity=0.8,
  draw opacity=1,
  text opacity=1,
  at={(0.09,0.6)},
  anchor=west,
  draw=lightgray204
},
tick align=outside,
tick pos=left,
x grid style={darkgray176},
xlabel={Number of Aisles},
xmin=3.75, xmax=31.25,
xtick style={color=black},
y grid style={darkgray176},
ylabel={Optimality Gap (\%)},
ymin=-0.552190932772972, ymax=62.7575110655829,
ytick style={color=black}
]
\addplot [semithick, steelblue31119180, dotted, mark=triangle*, mark size=3, mark options={solid,rotate=180}]
table {%
5 8.69948489009252
10 5.5599401626224
15 14.1127682176197
20 17.8499350424434
25 22.3499565254315
30 25.8544632436219
};
\addlegendentry{S-Shape}
\addplot [semithick, darkorange25512714, mark=diamond*, mark size=3, mark options={solid}]
table {%
5 59.8797973383849
10 59.570405550085
15 58.1518629481452
20 55.7111850823682
25 55.4514478281351
30 54.9927835643096
};
\addlegendentry{Return}
\addplot [semithick, forestgreen4416044, dashed, mark=pentagon*, mark size=3, mark options={solid}]
table {%
5 7.84359936901549
10 4.96973294747609
15 7.8707697075085
20 9.42865105256809
25 11.0396642134843
30 11.8694796856241
};
\addlegendentry{Composite}
\addplot [semithick, crimson2143940, dash pattern=on 1pt off 3pt on 3pt off 3pt, mark=square*, mark size=3, mark options={solid}]
table {%
5 17.4149017971706
10 18.3335834848761
15 12.9493940218501
20 9.93082563942591
25 7.68197723642206
30 6.0289404129117
};
\addlegendentry{Largest Gap}
\addplot [semithick, mediumpurple148103189, mark=triangle*, mark size=3, mark options={solid}]
table {%
5 5.28506139925509
10 2.32552279442502
15 2.57337646658845
20 2.83052382660003
25 2.67715067960409
30 2.36744333536411
};
\addlegendentry{Model (Standard)}
\addplot [semithick, sienna1408675, dotted, mark=*, mark size=3, mark options={solid}]
table {%
5 6.54246057859868
10 3.1645156687211
15 4.32553621233914
20 4.7597676113936
25 5.15576005224744
30 5.28439207627826
};
\addlegendentry{Model (Simplified)}
\end{axis}

\end{tikzpicture}
        }
        \label{fig:aisle_results}
    \end{subfigure}
    \begin{subfigure}[b]{0.45\textwidth}
    \centering
        \resizebox{\linewidth}{!}{
\begin{tikzpicture}

\definecolor{crimson2143940}{RGB}{214,39,40}
\definecolor{darkgray176}{RGB}{176,176,176}
\definecolor{darkorange25512714}{RGB}{255,127,14}
\definecolor{forestgreen4416044}{RGB}{44,160,44}
\definecolor{lightgray204}{RGB}{204,204,204}
\definecolor{mediumpurple148103189}{RGB}{148,103,189}
\definecolor{sienna1408675}{RGB}{140,86,75}
\definecolor{steelblue31119180}{RGB}{31,119,180}

\begin{axis}[
legend cell align={left},
legend style={
  fill opacity=0.8,
  draw opacity=1,
  text opacity=1,
  at={(0.91,0.6)},
  anchor=east,
  draw=lightgray204
},
tick align=outside,
tick pos=left,
x grid style={darkgray176},
xlabel={Number of Items},
xmin=27, xmax=93,
xtick style={color=black},
y grid style={darkgray176},
ylabel={Optimality Gap (\%)},
ymin=-0.232385733869069, ymax=62.0728741602936,
ytick style={color=black}
]
\addplot [semithick, steelblue31119180, dotted, mark=triangle*, mark size=3, mark options={solid,rotate=180}]
table {%
30 26.5277816343987
45 19.1056740485068
60 14.4300099648202
75 10.3648566568226
90 8.26046776364463
};
\addlegendentry{S-Shape}
\addplot [semithick, darkorange25512714, mark=diamond*, mark size=3, mark options={solid}]
table {%
30 54.7341078962261
45 56.533254353526
60 57.8899857664128
75 58.0664036843147
90 59.2408168923771
};
\addlegendentry{Return}
\addplot [semithick, forestgreen4416044, dashed, mark=pentagon*, mark size=3, mark options={solid}]
table {%
30 12.5894666257253
45 10.6702659865136
60 8.74497611356812
75 6.74802540893074
90 5.4321800116594
};
\addlegendentry{Composite}
\addplot [semithick, crimson2143940, dash pattern=on 1pt off 3pt on 3pt off 3pt, mark=square*, mark size=3, mark options={solid}]
table {%
30 6.76576646402729
45 9.50354907777938
60 12.0751928582013
75 14.5795039870171
90 17.3590064401886
};
\addlegendentry{Largest Gap}
\addplot [semithick, mediumpurple148103189, mark=triangle*, mark size=3, mark options={solid}]
table {%
30 3.02751211622457
45 3.13523492975172
60 3.20193618544351
75 3.08487731939678
90 2.59967153404741
};
\addlegendentry{Model (Standard)}
\addplot [semithick, sienna1408675, dotted, mark=*, mark size=3, mark options={solid}]
table {%
30 5.62011919252445
45 5.59712152185094
60 5.29498029098186
75 4.20932993562838
90 3.63880922532955
};
\addlegendentry{Model (Simplified)}
\end{axis}

\end{tikzpicture}
        }
        \label{fig:item_results}
    \end{subfigure}
\vspace{-0.5cm}
\caption{Average optimality gap of proposed model compared to common heuristics
        for problem classes grouped by number of aisles (left) and size of pick
        list (right).} 
\label{fig:results}
\vskip 0.1in
\end{figure}
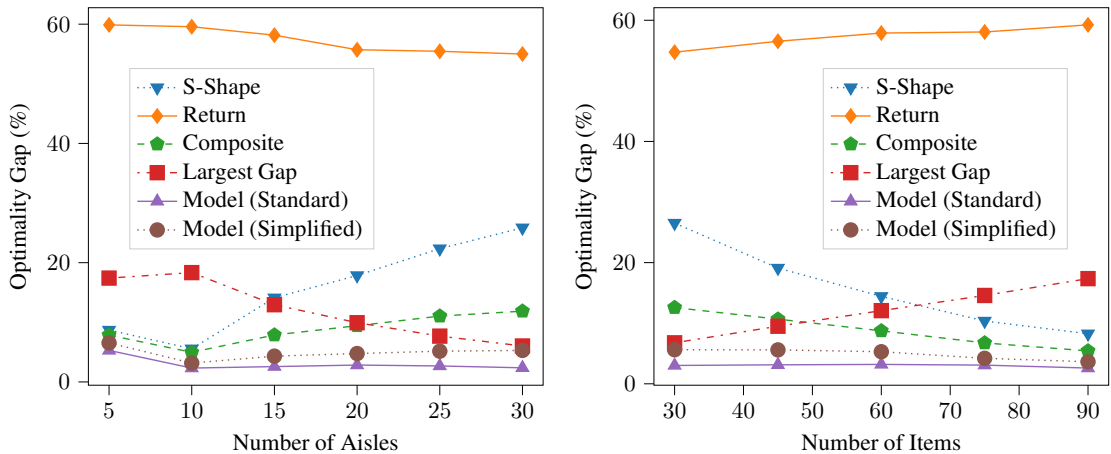


\section{Conclusion}
\label{conclusion}

This paper presented the problem of creating tour subgraphs for
picker routing in the form of a Markov Decision Process.
A neural network was then used to model this process using attention
mechanisms; with the parameters optimized using Reinforcement
Learning.
The ability of our approach to be customized for
additional constraints was demonstrated by modifications
that provide less complex routes, which is often
desirable within human-operated warehouses.
Numerical experiments found that the network outperformed
existing heuristics that are commonly used in practice
on a wide range of warehouse and problem sizes.

The results of this work show promise for further extensions
to related problems such as
dynamic order picking, warehouses with multiple
cross-aisles, and order batching,
representing even greater opportunities for
this work to offer contributions to the
optimization of warehouse operations.





\section*{Acknowledgements}

George was supported by an Australian Government Research Training
Program (RTP) Scholarship.


\section*{Impact Statement}

This study introduces advancements aimed at enhancing the order picking process in warehouses,
whether operated by humans or robots, through the application of machine learning.
The substantial societal impacts of this research are evident,
considering that warehousing operations incur significant costs within supply chains,
with a predominant portion attributed to the order picking process. 
Machine learning techniques, renowned for their speed and accuracy,
offer promising prospects for supplanting the prevailing heuristics in decision-making processes.
These heuristics, while currently dominant, often deviate from optimal solutions. 
The machine learning approach proposed in this study distinguishes itself by offering users
the flexibility to simplify perceived solution complexities,
which proves especially advantageous for human-operated warehouses.



\bibliographystyle{apalike}


\newpage
\appendix
\onecolumn
\section{Appendix}


\subsection{Partial Tour Subgraph States}
\label{sec:pts_states}

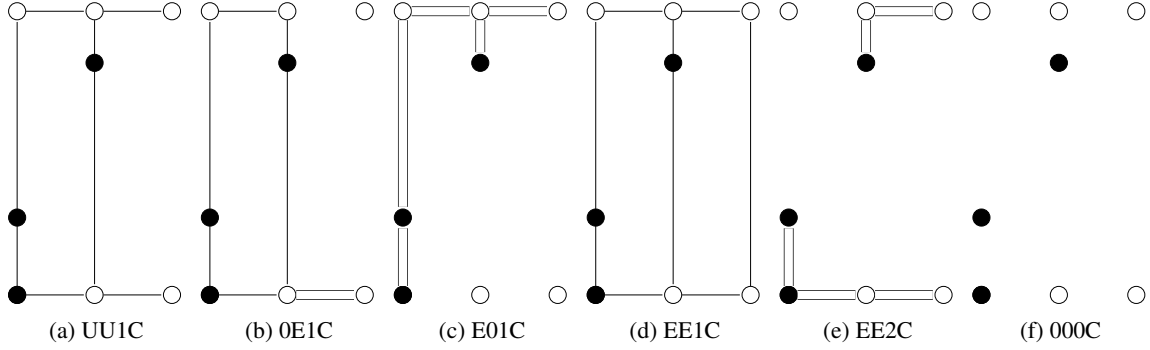
\begin{figure}[h]
\vskip 0.2in
\begin{center}
    \begin{subfigure}[b]{0.15\linewidth}
        \centering
        \resizebox{\linewidth}{!}{
        \begin{tikzpicture}[shorten >=1pt,draw=black!50]




    \foreach \name / \y in {0,...,2}{
        \node[shape=circle,draw=black, minimum size=3pt, fill=white] (B-\name) at (1.5 * \y, -0.25) {};
        \node[shape=circle,draw=black, minimum size=3pt, fill=white] (A-\name) at (1.5 * \y , 5.25) {};

        }


    \node[circle, draw=black, minimum size=3pt, fill=black] (depot) at (0, -0.25) {};

    \node[circle, draw=black, minimum size=3pt, fill=black] (item1) at (0, 1.25) {};

    \node[circle,draw=black, minimum size=3pt, fill=black] (item2) at (1.5, 4.25) {};








    \draw[black, thin] (depot) -- (A-0);
    \draw[black, thin] (A-0) -- (A-1);
    \draw[black, thin] (A-1) -- (B-1);
    \draw[black, thin] (B-1) -- (B-2);
    \draw[black, thin] (B-0) -- (B-1);
    \draw[black, thin] (A-1) -- (A-2);

\end{tikzpicture}
        }
        \caption{UU1C}
        \label{fig:uu1c}
    \end{subfigure}
    \begin{subfigure}[b]{0.15\linewidth}
        \centering
        \resizebox{\linewidth}{!}{
        \begin{tikzpicture}[shorten >=1pt,draw=black!50]




    \foreach \name / \y in {0,...,2}{
        \node[shape=circle,draw=black, minimum size=3pt, fill=white] (B-\name) at (1.5 * \y, -0.25) {};
        \node[shape=circle,draw=black, minimum size=3pt, fill=white] (A-\name) at (1.5 * \y , 5.25) {};

        }


    \node[circle, draw=black, minimum size=3pt, fill=black] (depot) at (0, -0.25) {};

    \node[circle, draw=black, minimum size=3pt, fill=black] (item1) at (0, 1.25) {};

    \node[circle,draw=black, minimum size=3pt, fill=black] (item2) at (1.5, 4.25) {};








    \draw[black, thin] (depot) -- (A-0);
    \draw[black, thin] (A-0) -- (A-1);
    \draw[black, thin] (A-1) -- (B-1);
    \draw[black, thin, double, double distance between line centers=5pt] (B-1) -- (B-2);
    \draw[black, thin] (B-0) -- (B-1);

\end{tikzpicture}
        }
        \caption{0E1C}
        \label{fig:0e1c}
    \end{subfigure}
    \begin{subfigure}[b]{0.15\linewidth}
        \centering
        \resizebox{\linewidth}{!}{
        \begin{tikzpicture}[shorten >=1pt,draw=black!50]




    \foreach \name / \y in {0,...,2}{
        \node[shape=circle,draw=black, minimum size=3pt, fill=white] (B-\name) at (1.5 * \y, -0.25) {};
        \node[shape=circle,draw=black, minimum size=3pt, fill=white] (A-\name) at (1.5 * \y , 5.25) {};

        }


    \node[circle, draw=black, minimum size=3pt, fill=black] (depot) at (0, -0.25) {};

    \node[circle, draw=black, minimum size=3pt, fill=black] (item1) at (0, 1.25) {};

    \node[circle,draw=black, minimum size=3pt, fill=black] (item2) at (1.5, 4.25) {};








    \draw[black, thin, double, double distance between line centers=5pt] (depot) -- (item1);
    \draw[black, thin, double, double distance between line centers=5pt] (A-0) -- (item1);
    \draw[black, thin, double, double distance between line centers=5pt] (A-0) -- (A-1);
    \draw[black, thin, double, double distance between line centers=5pt] (A-1) -- (item2);
    \draw[black, thin, double, double distance between line centers=5pt] (A-1) -- (A-2);

\end{tikzpicture}
        }
        \caption{E01C}
        \label{fig:e01c}
    \end{subfigure}
    \begin{subfigure}[b]{0.15\linewidth}
        \centering
        \resizebox{\linewidth}{!}{
        \begin{tikzpicture}[shorten >=1pt,draw=black!50]




    \foreach \name / \y in {0,...,2}{
        \node[shape=circle,draw=black, minimum size=3pt, fill=white] (B-\name) at (1.5 * \y, -0.25) {};
        \node[shape=circle,draw=black, minimum size=3pt, fill=white] (A-\name) at (1.5 * \y , 5.25) {};

        }


    \node[circle, draw=black, minimum size=3pt, fill=black] (depot) at (0, -0.25) {};

    \node[circle, draw=black, minimum size=3pt, fill=black] (item1) at (0, 1.25) {};

    \node[circle,draw=black, minimum size=3pt, fill=black] (item2) at (1.5, 4.25) {};








    \draw[black, thin] (depot) -- (A-0);
    \draw[black, thin] (A-0) -- (A-1);
    \draw[black, thin] (A-1) -- (B-1);
    \draw[black, thin] (B-1) -- (B-2);
    \draw[black, thin] (B-0) -- (B-1);
    \draw[black, thin] (A-1) -- (A-2);
    \draw[black, thin] (B-2) -- (A-2);

\end{tikzpicture}
        }
        \caption{EE1C}
        \label{fig:ee1c}
    \end{subfigure}
        \begin{subfigure}[b]{0.15\linewidth}
        \centering
        \resizebox{\linewidth}{!}{
        \begin{tikzpicture}[shorten >=1pt,draw=black!50]




    \foreach \name / \y in {0,...,2}{
        \node[shape=circle,draw=black, minimum size=3pt, fill=white] (B-\name) at (1.5 * \y, -0.25) {};
        \node[shape=circle,draw=black, minimum size=3pt, fill=white] (A-\name) at (1.5 * \y , 5.25) {};

        }


    \node[circle, draw=black, minimum size=3pt, fill=black] (depot) at (0, -0.25) {};

    \node[circle, draw=black, minimum size=3pt, fill=black] (item1) at (0, 1.25) {};

    \node[circle,draw=black, minimum size=3pt, fill=black] (item2) at (1.5, 4.25) {};








    \draw[black, thin, double, double distance between line centers=5pt] (depot) -- (item1);
    \draw[black, thin, double, double distance between line centers=5pt] (A-1) -- (item2);
    \draw[black, thin, double, double distance between line centers=5pt] (B-1) -- (B-2);
    \draw[black, thin, double, double distance between line centers=5pt] (B-0) -- (B-1);
    \draw[black, thin, double, double distance between line centers=5pt] (A-1) -- (A-2);

\end{tikzpicture}
        }
        \caption{EE2C}
        \label{fig:ee2c}
    \end{subfigure}
        \begin{subfigure}[b]{0.15\linewidth}
        \centering
        \resizebox{\linewidth}{!}{
        \begin{tikzpicture}[shorten >=1pt,draw=black!50]




    \foreach \name / \y in {0,...,2}{
        \node[shape=circle,draw=black, minimum size=3pt, fill=white] (B-\name) at (1.5 * \y, -0.25) {};
        \node[shape=circle,draw=black, minimum size=3pt, fill=white] (A-\name) at (1.5 * \y , 5.25) {};

        }


    \node[circle, draw=black, minimum size=3pt, fill=black] (depot) at (0, -0.25) {};

    \node[circle, draw=black, minimum size=3pt, fill=black] (item1) at (0, 1.25) {};

    \node[circle,draw=black, minimum size=3pt, fill=black] (item2) at (1.5, 4.25) {};









\end{tikzpicture}
        }
        \caption{000C}
        \label{fig:000C}
    \end{subfigure}
\caption{Examples of Partial Tour Subgraphs for each state.} 
\label{fig:equivalence_states}
\end{center}
\vskip -0.2in
\end{figure}


\subsection{Aisle Encoding}
\label{sec:encoding}

The ``aisle encodings'' are based on the
``position encodings'' in \citet{vaswani2017attention}, 
defined by:
\begin{align}
    & AE(i, 2j) = \sin(i / 10000^{2j/d_{h}}), \\
    & AE(i, 2j+1) = \cos(i / 10000^{2j/d_{h}}),
\end{align}
where $i$ is the position in the sequence and $j$ is the element.


\subsection{Encoder Layers}
\label{sec:encoder}

Following on from the Transformer encoder in \citet{vaswani2017attention},
each attention layer consists of two sub-layers;
a Multi-Head Attention (MHA) layer and
an aisle-wise fully connected feedforward (FF) layer.
Each sublayer adds a skip connection \cite{he2016deep}
and layer normalization \cite{ba2016layer}.
There are $N$ layers in total,
each with its own parameters.

\textbf{Attention.}
\citet{vaswani2017attention}
computes the key $\mathbf{k}_i \in \mathbb{R}^{d_k}$,
value $\mathbf{v}_i \in \mathbb{R}^{d_v}$ and
query $\mathbf{q}_i \in \mathbb{R}^{d_k}$
for each input embedding $\mathbf{h}_i$ by:
\begin{equation}
    \mathbf{q}_i = W^Q \mathbf{h}_i, \ 
    \mathbf{k}_i = W^K \mathbf{h}_i, \ 
    \mathbf{v}_i = W^V \mathbf{h}_i,
\end{equation}
$W^Q$, $W^K$ (both of size ($d_k \times d_h$) and
$W^V$ ($d_v \times d_h$) are learnable parameters.

With an attention mask $M \in \mathbb{B}^{n \times n}$,
the compatibility $u_{ij} \in \mathbb{R}$ of query $\mathbf{q}_i$
to key $\mathbf{k}_j$ is then given by the scaled
dot-product:
\begin{equation}
    u_{ij} = 
    \begin{cases}
        \frac{\mathbf{q}^T_i \mathbf{k}_j}
        {\sqrt{d_k}} & \text{if} \quad M_{ij} = \text{False}, \\
        -\infty & \text{otherwise}. \\
    \end{cases}
\end{equation}

The attention weights $a_{ij} \in [0,1]$ are then calculated
using a softmax:
\begin{equation}
    a_{ij} = \frac{e^{u_{ij}}}{\sum_{j'} e^{u_{ij'}}}.
\end{equation}

The output vector $\mathbf{h}'_i$ is then the weighted sum
of the values $\mathbf{v}_j$:
\begin{equation}
    \mathbf{h}'_i = \sum_j a_{ij} \mathbf{v}_j.
\end{equation}

The Transformer encoder utilizes self-attention,
which means that all keys,
values, and queries come from the same place.
These are the aisle embeddings in the first layer
and the output of the previous layer for all subsequent layers.

\textbf{Multi-head Attention.}
It has been found to be beneficial to have multiple heads of
attention \cite{vaswani2017attention, velivckovic2017graph}.
In our application,
this allows each aisle to simultaneously attend to different
information from other aisles and
is done by linearly projecting the keys, 
values, and queries multiple times and performing the
attention function on each of the projections. 
The outputs are then concatenated to give the final values.
This allows the model to attend to different information simultaneously,
which has been found to be beneficial.

The queries, keys and values are projected $H$ times to 
dimension $d_k = d_v = d_h/H$ and
attention is performed on all projections before
one final projection:
\begin{align}
    & \text{head}_i = \text{Attention}(QW^Q_i, KW^K_i, VW^V_i), \\
    & \text{MHA}(Q, K, V) 
    = \text{Concat}(\text{head}_1,...,\text{head}_H)W^O,
\end{align}
where $W^Q_i \in \mathbb{R}^{d_h \times d_k}$,
$W^K_i \in \mathbb{R}^{d_h \times d_k}$,
$W^V_i \in \mathbb{R}^{d_h \times d_v}$ and
$W^O \in \mathbb{R}^{d_h \times d_h}$.

\textbf{Feed-Forward.}
Two fully connected linear transformations 
applied to each aisle position separately 
with a ReLU activation in between:
\begin{equation}
    \text{FF}(\hat{\mathbf{h}}_i) 
    = \text{ReLU}
    (\hat{\mathbf{h}}_i W^{ff}_1 + b^{ff}_1) W^{ff}_2 + b^{ff}_2.
\end{equation}
The linear transformations are the same for each input position
but use different parameters for each layer.
Both the input and output have dimensions $d_h$
and the dimension of the hidden layer is $d_{ff} = 4d_h$.


\subsection{REINFORCE Algorithm}

\begin{algorithm} 
    \caption{REINFORCE with Rollout Baseline}
    \label{reinforce}
\begin{algorithmic}
    \STATE \textbf{Input:} number of Epochs E, 
    steps per epoch T,
    batch size B,
    significance $\alpha$
    \STATE Initialize $\theta$, $\theta^{BL} \gets \theta$
    \FOR{epoch $= 1,...,$E}
        \FOR{step $= 1,...,$T}
            \STATE $s_i \gets$ RandomInstance() $\forall i \in \{1,...,\text{B}\}$
            \STATE $\pi_i \gets$ SampleRollout($s_i, p_\theta$) 
            $\forall i \in \{1,...,\text{B}\}$
            \STATE $\pi^{BL}_i \gets$ GreedyRollout($s_i, p_{\theta^{BL}}$) 
            $\forall i \in \{1,...,\text{B}\}$
            \STATE $\nabla \mathcal{L} \gets \sum_{i=1}^B 
            \frac{(L(\pi_i) - L(\pi^{BL}_i))}{L(\pi^{BL}_i)}
            \nabla_\theta \log p_\theta (\pi_i)$
            \STATE $\theta \gets \text{Adam}(\theta, \nabla \mathcal{L})$
        \ENDFOR
    \IF{OneSidedPairedTTest($p_\theta, p_{\theta^{BL}}$) $< \alpha$}
        \STATE $\theta^{BL} \gets \theta$
    \ENDIF
    \ENDFOR
\end{algorithmic}
\end{algorithm}
\label{sec:reinforce}


\newpage
\subsection{Experiment Results}
\label{sec:results_table}

\begin{table}[h]
\centering
\caption {Average optimality gap (\%) of proposed and heuristic methods for all problem classes.}
\vskip 0.1in
\begin{tabular}{ccccccccc}
\toprule
 Aisles & Items & S-Shape & Return & Composite & Largest Gap & Model (Standard) & Model (Simplified) \\
\midrule
5 & 30 &      13.86 &      57.89 &      10.66 &      11.75 &       4.34 &       6.31 \\
5 & 45 &       9.40 &      60.07 &       8.99 &      16.03 &       5.22 &       7.40 \\
5 & 60 &       8.00 &      61.02 &       7.77 &      18.73 &       6.05 &       7.28 \\
5 & 75 &       6.48 &      60.48 &       6.18 &      19.68 &       5.61 &       6.04 \\
5 & 90 &       5.76 &      59.94 &       5.62 &      20.88 &       5.21 &       5.68 \\
10 & 30 &      16.51 &      56.03 &      11.97 &       8.99 &       3.40 &       5.69 \\
10 & 45 &       7.13 &      57.72 &       6.63 &      14.11 &       3.17 &       3.99 \\
10 & 60 &       3.01 &      58.18 &       3.76 &      18.82 &       2.39 &       3.50 \\
10 & 75 &       0.98 &      61.74 &       1.93 &      22.43 &       2.03 &       1.88 \\
10 & 90 &       0.17 &      64.19 &       0.56 &      27.32 &       0.64 &       0.76 \\
15 & 30 &      27.47 &      55.95 &      12.40 &       6.76 &       3.15 &       6.21 \\
15 & 45 &      16.82 &      57.23 &       9.97 &       9.56 &       2.66 &       5.00 \\
15 & 60 &      11.63 &      58.52 &       7.71 &      13.10 &       2.40 &       3.86 \\
15 & 75 &       7.88 &      58.79 &       4.79 &      15.94 &       2.39 &       3.17 \\
15 & 90 &       6.76 &      60.26 &       4.48 &      19.39 &       2.27 &       3.39 \\
20 & 30 &      30.34 &      53.77 &      12.07 &       5.23 &       3.10 &       5.17 \\
20 & 45 &      24.00 &      56.56 &      12.61 &       7.07 &       3.06 &       6.11 \\
20 & 60 &      16.47 &      56.40 &       9.44 &       9.51 &       2.72 &       5.25 \\
20 & 75 &      10.67 &      55.00 &       7.37 &      12.45 &       2.88 &       3.98 \\
20 & 90 &       7.77 &      56.83 &       5.65 &      15.39 &       2.38 &       3.30 \\
25 & 30 &      35.56 &      55.10 &      14.70 &       4.74 &       2.68 &       5.58 \\
25 & 45 &      26.72 &      53.27 &      12.42 &       6.10 &       2.64 &       5.63 \\
25 & 60 &      20.91 &      55.13 &      11.32 &       6.71 &       2.87 &       5.86 \\
25 & 75 &      16.05 &      56.90 &       9.43 &       9.51 &       2.72 &       4.88 \\
25 & 90 &      12.51 &      56.85 &       7.32 &      11.35 &       2.47 &       3.82 \\
30 & 30 &      35.44 &      49.67 &      13.72 &       3.12 &       1.50 &       4.75 \\
30 & 45 &      30.57 &      54.35 &      13.38 &       4.15 &       2.06 &       5.46 \\
30 & 60 &      26.55 &      58.09 &      12.47 &       5.58 &       2.78 &       6.02 \\
30 & 75 &      20.12 &      55.48 &      10.80 &       7.47 &       2.88 &       5.30 \\
30 & 90 &      16.59 &      57.37 &       8.98 &       9.82 &       2.62 &       4.89 \\
\bottomrule
\end{tabular}
\label{tab:results_table}
\end{table}


\end{document}